\documentclass[oneside,english]{article}

\usepackage[utf8]{inputenc}
\usepackage{hyperref}
\usepackage{url}
\usepackage{algorithm,algpseudocode}
\usepackage{mathtools}
\usepackage{amsmath,amssymb,amsthm}
\usepackage{bbm}
\usepackage{graphicx}
\usepackage{mathptmx}
\usepackage{todonotes}
\usepackage{enumitem}
\usepackage{esint}
\usepackage{booktabs}
\usepackage{multirow}

\usepackage{titlesec}
\titleformat{\section}
	{\normalfont\Large\bfseries\filcenter}{\thesection.}{1 ex}{}
\titleformat{\subsection}
	{\normalfont\normalsize\bfseries}{\thesubsection.}{1 ex}{}
\titleformat{\subsubsection}[runin]
	{\normalfont\normalsize\bfseries\filcenter}{\thesubsubsection.}{1 ex}{}

\usepackage[charter]{mathdesign}
\usepackage[mathcal]{eucal}

\usepackage[margin=1.2 in]{geometry}

\usepackage[square,numbers,sort&compress]{natbib}

\usepackage{thmtools}

\makeatletter
\providecommand*{\diff}%
        {\@ifnextchar^{\DIfF}{\DIfF^{}}}
\def\DIfF^#1{%
        \mathop{\mathrm{\mathstrut d}}%
                \nolimits^{#1}\gobblespace
}
\def\gobblespace{%
        \futurelet\diffarg\opspace}
\def\opspace{%
        \let\DiffSpace\!%
        \ifx\diffarg(%
                \let\DiffSpace\relax
        \else
                \ifx\diffarg\[%
                        \let\DiffSpace\relax
                \else
                        \ifx\diffarg\{%
                                \let\DiffSpace\relax
                        \fi\fi\fi\DiffSpace}
\makeatother

\makeatletter

\numberwithin{equation}{section}
\numberwithin{figure}{section}
\theoremstyle{plain}

  \theoremstyle{definition}
  
  \theoremstyle{definition}
  
  \theoremstyle{plain}
  
  \theoremstyle{plain}
  
  \theoremstyle{remark}
  
  \theoremstyle{plain}

\makeatother

\usepackage{babel}
  \providecommand{\corollaryname}{Corollary}
  \providecommand{\definitionname}{Definition}
  \providecommand{\examplename}{Example}
  \providecommand{\lemmaname}{Lemma}
  \providecommand{\remarkname}{Remark}
  \providecommand{\theoremname}{Theorem}
  \providecommand{\propositionname}{Proposition}

\usepackage{authblk}
\usepackage{multirow}
\usepackage{makecell}
\usepackage{longtable}

\usepackage{graphicx}
	\graphicspath{{results/images/}}

\usepackage{multibib}
\newcites{nips}{References – Reviewed Papers}

\title{NIPS - Not Even Wrong?\\ {\large A Systematic Review of Empirically Complete Demonstrations of Algorithmic Effectiveness in the Machine Learning and Artificial Intelligence Literature} }

\author[1,2]{
Franz J.~Kir\'{a}ly
\thanks{\url{fkiraly@turing.ac.uk}}
}

\author[1,3,4]{
Bilal Mateen
\thanks{\url{bmateen@turing.ac.uk}}
}

\author[1,2]{Raphael Sonabend
\thanks{\url{raphael.sonabend.15@ucl.ac.uk}}
}

\affil[1]{The Alan Turing Institute,\newline
Kings Cross,
London NW1 2DB, United Kingdom
}

\affil[2]{
Department of Statistical Science,
University College London,\newline
Gower Street,
London WC1E 6BT, United Kingdom
}

\affil[3]{Warwick Medical School,
University of Warwick,\newline
Coventry CV4 7AL, United Kingdom
}

\affil[4]{Kings College Hospital,
Denmark Hill,\newline
London SE5 9RS,  United Kingdom
}

\begin{document}

\maketitle

\begin{abstract}
Objective: To determine the completeness of argumentative steps necessary to conclude effectiveness of an algorithm in a sample of current ML/AI supervised learning literature.

Data Sources: Papers published in the `Neural Information Processing Systems’ (NeurIPS, n\'ee NIPS) journal where the official record showed a 2017 year of publication.

Eligibility Criteria: Studies reporting a (semi-)supervised model, or pre-processing fused with (semi-)supervised models for tabular data.

Study Appraisal: Three reviewers applied the assessment criteria to determine argumentative completeness. The criteria were split into three groups, including: experiments (e.g real and/or synthetic data), baselines (e.g uninformed and/or state-of-art) and quantitative comparison (e.g. performance quantifiers with confidence intervals and formal comparison of the algorithm against baselines).

Results: Of the 121 eligible manuscripts (from the sample of 679 abstracts), 99\% used real-world data and 29\% used synthetic data. 91\% of manuscripts did not report an uninformed baseline and 55\% reported a state-of-art baseline. 32\% reported confidence intervals for performance but none provided references or exposition for how these were calculated. 3\% reported formal comparisons.

Limitations: The use of one journal as the primary information source may not be representative of all ML/AI literature. However, the NeurIPS conference is recognised to be amongst the top tier concerning ML/AI studies, so it is reasonable to consider its corpus to be representative of high-quality research.

Conclusion: Using the 2017 sample of the NeurIPS supervised learning corpus as an indicator for the quality and trustworthiness of current ML/AI research, it appears that complete argumentative chains in demonstrations of algorithmic effectiveness are rare.
\end{abstract}

\newpage
\section{Machine learning and AI: data ... science?}
\label{sec:intro}

With the data science revolution in full motion, the real world reach and scope of data scientific methodology is extending rapidly - particularly driven by methodology commonly subsumed under the ``machine learning'', ``advanced analytics'', or ``artificial intelligence'' labels (which we will abbreviate to ML/AI). As industrial sectors, society, and policy makers face a deluge of an ever-growing collection of ML/AI algorithms, the natural question poses itself: to what extent is a novel technology effective? And, to what extent may its promises be trusted?

Systematic reviews are common in fields of science with similar societal and reach and impact, such as modern, evidence-based medicine~\cite{Sackett71}. They provide a robust, unbiased, summary of the evidence on a specific topic, with which to inform discussion, and facilitate decision-making. In this systematic review, we attempt to paint a representative, cross-sectional picture of how effectiveness is empirically evidenced in recently published ML/AI work.

\begin{enumerate}
\itemsep-0.2em
\item[(i)] As the representative corpus, we consider publications accepted and published at NIPS. We consider the annual NIPS conference to be amoungst the top tier conferences in machine learning and artificial intelligence, and the one with highest visibility towards society, industry, and the public sector. We hence consider the NIPS corpus as representative of both the highest quality and most extensive reception, when it comes to ML/AI research.

\item[(ii)] As the key criterion, we look for argumentative completeness in drawing the scientific conclusion that ``the new method does something useful''. We investigate for the presence of the necessary arguments defining a scientifically testable claim. The opposite would be results which are ``not even wrong'' in the words of Wolfgang Pauli~\cite{peierls19001958}, i.e., pseudo-scientific. We explicitly avoid discussion and assessment of the technical (e.g., statistical) methodology by which the arguments are made, to keep the bar low. A full argumentative chain can still fail from faulty mathematics or statistics, but if the scientific argument is incomplete, no level of mathematical sophistication can fix it.

\item[(iii)] We restrict ourselves to innovations in the field of supervised learning, one of the oldest and most popular sub-fields of ML/AI. There are two reasons for this choice: first, we believe that supervised learning may be considered an ``indicator'' sub-field of ML/AI research, in which we would expect standards in reporting and scientific argumentation to be most refined, when compared to other sub-fields. Second, in supervised learning, there is a generally accepted consensus on the key technical criteria evidencing a new supervised learning method as useful - namely, that it predicts with lower error than state-of-art baselines, or predicts with similar accuracy but with some other improved metric (e.g. run-time) - which facilitates systematic reviewing.
\end{enumerate}

Our main aim is to identify, using the indicator corpus, whether in contemporary ML/AI literature, the necessary argumentative steps to conclude effectiveness of an algorithm have been undertaken. We would like to emphasize that our focus lies on \emph{validity of argumentation}, rather than the \emph{statistical methodology} utilised to make the argument (such as in the review contained in~\cite{demsar2006statistical}). Correct statistical methodology is an important facet of, but ultimately only a part of the full empirical argument to conclude that a novel algorithm represents an improvement on the state-of-art method.

It is worth noting at this point that in both the ML/AI and medical statistics literature, there is growing criticism of null-hypothesis significance testing (NHST), also known as frequentist hypothesis testing, as the default mechanism for demonstrations of effectiveness, as it is often either, incorrectly interpreted, inappropriately applied, or misguidedly utilised as the gatekeeper for publication~\cite{szucs2017hypothesis, demsar2008appropriateness}.
Whilst these criticisms are ultimately reasonable, they pertain to misuse and mal-incentives in the context of publication mechanisms; they are \emph{not} in any form a criticism of quantitative comparison in itself as a necessary part of the argumentative chain (as it is occasionally falsely asserted). Since an argument that does not state how test the claim is, by definition, not testable, hence unscientific.

The rest of the paper is structured as follows: Section~2 describes the review methodology. Section~3 reports statistical summaries and key findings. Section~4 provides a discussion of our results. The appendix contains a full review documentation.

\section{Methodology}

\subsection{Information Sources}
The initial dataset consisted of all the manuscripts published in the academic journal ‘Neural Information Processing Systems’ (NeurIPS, née NIPS), where the official record showed the year of publication as 2017. No additional search constraints were applied. The python code for scraping this dataset from the web is available in the supplementary material.

\subsection{Study Eligibility}
\paragraph{Reviewers} The reviewer pool consisted of three reviewers; one academic with over ‘15’ years of experience in machine learning and data science, referred to as ``senior reviewer'' in the below, and 2 academics with post-graduate experience or equivalent in machine learning and data science, referred to as ``junior reviewers'' below.

\paragraph{Initial Screening Protocol} Each manuscript was assigned two screening reviewers, selected uniformly at random and subject to the constraint that each reviewer assessed the same number of manuscripts (plus-minus one). The assigned screening reviewers determined eligibility of the manuscript for full-text review, solely based on the title and published abstract. In situations where the two screening reviewers disagreed, the remaining third reviewer (from the three-reviewer-pool) assessed the abstract independently. A manuscript was always retained for full-text assessment, unless the threshold for exclusion was reached. The exclusion threshold was considered reached when the senior reviewer and at least one of the junior reviewers were in agreement to exclude. That is, the manuscript was retained for full-text assessment when either: both of the junior reviewers voted to retain, or the senior reviewer and at least one of the junior reviewers voted to retain.

\paragraph{Inclusion Criteria} The inclusion criteria at the initial screening stage was all abstracts which reported a new supervised or semi-supervised prediction algorithm, or pre-processing method fused with an established supervised/semi-supervised prediction algorithm.

\paragraph{Exclusion} All abstracts reporting only algorithms for problems other than supervised learning (e.g., unsupervised learning, reinforcement learning), or only theoretical results for established supervised methods, were excluded. Abstracts which implicitly or explicitly reported only methodology for non-tabular data or non-exchangeable generative processes (e.g., time series), were also excluded.

\subsection{Selecting the Assessment Criteria}
\label{Sec:methods.selQ}

The main purpose of the review is to check the completeness of the argumentative chain evidencing the proposed supervised method's effectiveness (``usefulness''). In reviewing, we explicitly restrict ourselves to the reporting of the argumentative chain, and exclude checks of technical, mathematical, statistical, algorithmical, or implementation correctness.

We consider an argumentative chain evidencing effectiveness complete, if it includes:

\begin{enumerate}
\itemsep-0.2em
\item[(a)] One or multiple experiments which test the supervised method against synthetic or real world data, optimally both. Usually, this will include observations of the estimated predictive performance using a variety of metrics. Without empirical observation of the performance, no empirical claim may be made.

\item[(b)] The use of suitable comparative baselines in the same experiment. This may include state-of-art competitors, and should include (directly or indirectly) some uninformed surrogate for ``guessing''. Without the first, no claim of outperforming the state-of-art can be made. Without the second, no claim of outperforming a random guess can be made.

\item[(c)] A formal quantitative assessment of performance, and a formal comparison to the comparative baselines. Such an assessment may consist of reporting confidence/credibility intervals (CI), or results of a frequentist hypothesis test. Without such an assessment a claim of effectiveness can not be made, since it has not been ruled out (subject to the usual caveats) that the observed differences in performance are not due to random fluctuation.
\end{enumerate}

As stated above, the review questions are \emph{not} meant to check whether the arguments are made correctly on a technical level, e.g., whether experiments are set up properly, performances and CI are estimated in a statistically sensible way, or whether hypothesis tests used are reasonable.\\
However, the review questions \emph{will} check whether all parts of the argumentative chain are present, and whether, from the publication text, it can be in-principle understood what exactly has been done. I.e., how were baselines selected, how were CI and test significances calculated. If this is not stated, then the reported results cannot undergo truly rigorous peer review, because whilst there are some good practice recommendations (e.g.,~\cite{demsar2006statistical}), there is no one consensus on how this analysis should be undertaken.

We describe the reviewer protocol and review items in more detail below.

\subsection{Reviewer protocol and review items}
\label{Sec:methods.protocol}

\paragraph{Reviewer protocol} The manuscripts retained for full-text assessment following the initial screening by abstract were again assigned to two reviewers selected uniformly at random, subject to the constraint that each reviewer assessed the same number of manuscripts (plus-minus one). Reviewers applied the criteria described below independently. Where there was disagreement on at least one review item, the third reviewer assessed the manuscript independently, again the majority outcome was recorded. In cases of disagreement between all three reviewers, these manuscripts were excluded from the rest and were discussed separately.
	
\begin{figure}[H]
\centering
\includegraphics[scale=1]{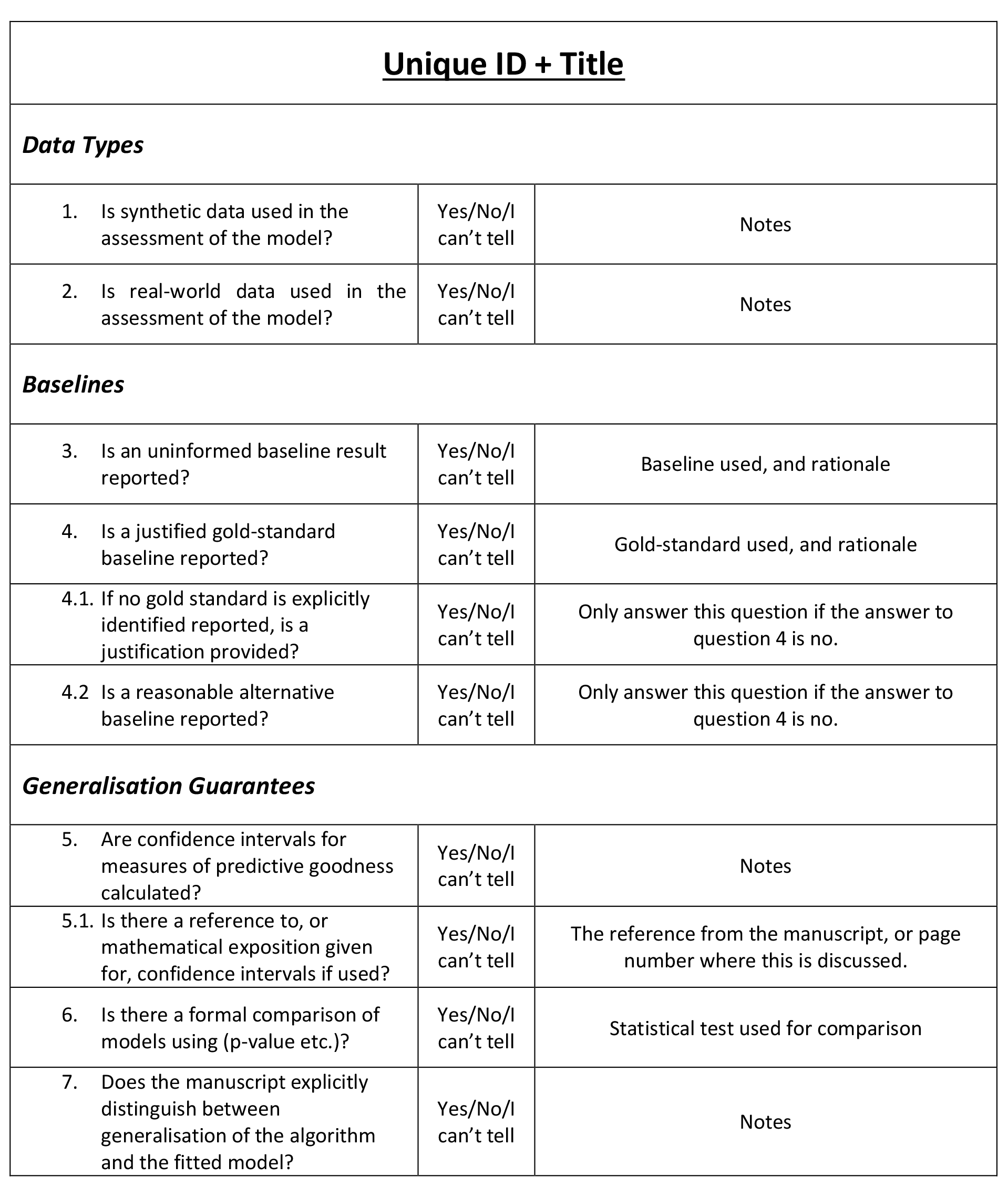}
\caption{Question sheet used for reviewing}
\label{fig:questions}
\end{figure}

\paragraph{Assessment domains.} The assessment criteria were split into three domains aligning with the key argumentative requirements identified in Section~\ref{Sec:methods.selQ}: reporting of experiments, baselines, and quantitative comparison. The review items are displayed in Figure~\ref{fig:questions}.
\\

{\bf Experiments.} This consists of two items (Fig. \ref{fig:questions}), querying whether a comparison experiment was conducted on (1) synthetic data, and/or (2) real world data. Synthetic data may be used to empirically test whether the proposed methodology works under the assumptions it is constructed from, or it can be mathematically shown to work under pre-specified constraints. Real world data may be used to empirically test whether the proposed methodology performs well on data from the real world. We consider at least one of these to be necessary for an empirical usefulness argument. While having both synthetic and a real world experiments provides a stronger argument as per the above discussion, we do not require reporting of both as necessary.\\

{\bf Baselines.} This is split into 2 items: (3) is a naive/uninformed baseline compared to, and (4) is a state-of-art baseline compared to. A negative response to item (4) leads to a series of additional items (Fig. \ref{fig:questions}): (4.1) whether there is an explanation for why none is compared to - e.g., being the first proposed method for a specific setting; and (4.2) whether a reasonable alternative to state-of-art comparison is present - e.g., reporting literature performance of a method whose code is difficult to obtain. We consider reporting of a (3) naive/uninformed baseline as necessary in making an empirical usefulness argument, at all. We consider reporting of (4) state-of-art baselines as necessary for making an argument evidencing out-performance of the state-of-art.\\

{\bf Quantitative comparisons.} This is split into 3 items: item (5) asks whether performance quantifiers with confidence intervals are reported. If reported, it is checked in an additional item (5.1) whether a literature reference for confidence interval computation is given, and/or whether the manuscript itself contains a mathematical exposition on how confidence intervals are computed. Item (6) checks whether the manuscript reports a formal comparison quantifier, such as for example results of a frequentist hypothesis test, or Bayesian credibility intervals. Item (7) asks for reviewer's judgement whether the authors of the manuscript have precisely outlined the situation for which the reported empirical results generalize to - e.g., whether they are a guarantee for performance of the fitted models, or the algorithms' performance when re-fitted, or whether there is no generalisation guarantee claimed. In line with the best practice to report effect size, significance quantifier (frequentist or not), and domain of (internal/external) validity, we would consider all items reported as necessary for an empirical usefulness argument.

\subsection{Analysis of Reviewer Agreement}

Post-hoc review data analysis was undertaken to examine reviewer agreement. This was conducted in two parts: (1) analysis of intra-manuscript agreement, i.e., the percentage of items pertainingtoa manuscript that pairs of reviewers agreed on. (2) analysis of intra-item agreement, i.e., the percentage of papers that pairs of reviewers agreed on for a given item.\\
In all plots and tables, `S' refers to the senior reviewer and `J1,J2' refer to the two junior reviewers. There are three possible pairs of reviewers, which are denoted as follows in plots and tables: `SJ1' is the pair of senior reviewer and junior reviewer one; `SJ1' is the pair of senior reviewer and junior reviewer two; `J1J2' is the pair of both junior reviewers.\\

Following \cite{fleiss2003statistical}, reviewer agreement was quantified by Cohen's Kappa and Fleiss' Kappa, both reported for all three possible pairs of reviewers. Confidence intervalsfor Cohen's Kappa are computed as described in~\cite{fleiss1969}. Multi-reviewer Kappas are not computed since in the first reviewing phase, all papers were reviewed by exactly two reviewers, and none by three (see first paragraph of Section~\ref{Sec:methods.protocol}). To quantify significance of a pairwise reviewer (dis)agreement, we use the two-sided frequentist hypothesis test for inexact Fleiss' kappa with the null hypothesis that overall rater agreement is due to chance\footnote{Null hypothesis is two-sided Kappa = 0, i.e., there is no association between review items; as implemented in the \texttt{kappam.fleiss} function of the \texttt{irr} package version 0.84}, so that a small p-value plausibly indicates genuine agreement or disagreement between at least two raters. We further follow the guidelines of \cite{fleiss2003statistical} to characterise reviewer agreement as measured by ranges of the Kappa statistic as: >0.75 - excellent, 0.40-0.75 - fair to good, <0.40 - poor.

\newpage
\section{Results}

\begin{figure}[h!]
\centering
\includegraphics[scale=0.5]{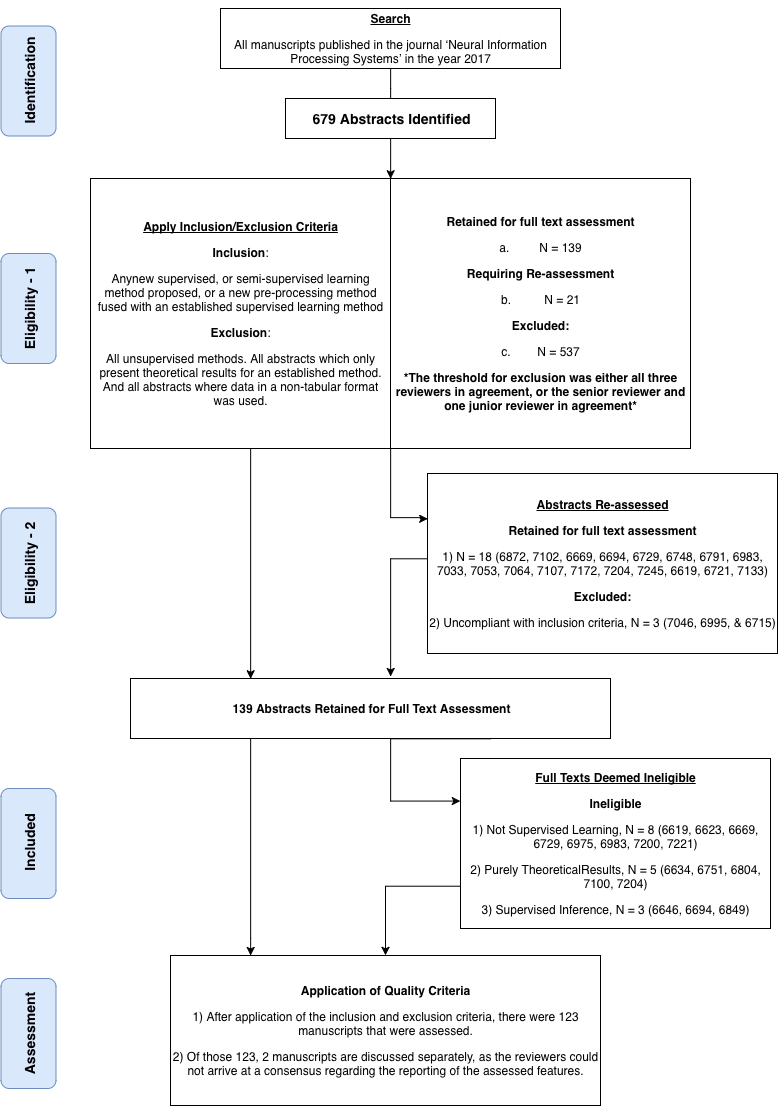}
\caption{PRISMA diagram of selection and reviewing process}
\label{fig:inclusion}
\end{figure}

A full PRISMA diagram \cite{prisma2009} of the manuscript selection process with the relevant inclusion/exclusion criteria is presented in Figure~\ref{fig:inclusion}. In summary, 139 manuscripts were retained for full-text assessment from the initial pool of 679 abstracts. 16 were excluded upon manuscript review as they did not meet the eligibility criteria, resulting in 123 manuscripts undergoing assessment using the aforementioned criteria. Two of these 123 manuscripts were excluded in the analyses below, and will be discussed separately, as the reviewers were unable to arrive at a conclusive consensus regarding reporting of the assessed features.

The consensus results for the final pool of 121 manuscripts are discussed in three parts (Table \ref{tab:conresults}), in the first we analyse the aggregated reviewer's responses for all papers, in the second we analyse the disagreement between the different reviewer pairings and in the third we discuss the two papers in which no consensus could be reached.

\begin{longtable}{p{0.6cm}|p{1cm}|p{0.7cm}|p{0.7cm}|p{1cm}|p{1.4cm}|p{1cm}|p{1cm}|p{0.3cm}|p{1cm}|p{1cm}|p{1cm}}
ID	&	Citation	&	Syn-thetic	&	Real-World	&	Unin-formed	&	Gold Standard	&	Explan-ation 	&	Alter-native	&	CIs &	CI Reference	& Compa-rison	&	General-isation	\\
\hline
6609	&	\citenips{NIPS2017_6609}	&	N	&	Y	&	N	&	Y	&	NA	&	NA	&	N	&	NA	&	N	&	N	\\
6614	&	\citenips{NIPS2017_6614}	&	N	&	Y	&	N	&	Y	&	NA	&	NA	&	N	&	NA	&	N	&	N	\\
6621	&	\citenips{NIPS2017_6621}	&	N	&	Y	&	Y	&	Y	&	NA	&	NA	&	Y	&	N	&	N	&	Y	\\
6624	&	\citenips{NIPS2017_6624}	&	Y	&	Y	&	N	&	Y	&	NA	&	NA	&	Y	&	N	&	N	&	Y	\\
6627	&	\citenips{NIPS2017_6627}	&	Y	&	Y	&	N	&	N	&	Y	&	Y	&	N	&	NA	&	N	&	N	\\
6637	&	\citenips{NIPS2017_6637}	&	Y	&	Y	&	N	&	Y	&	NA	&	NA	&	Y	&	N	&	N	&	N	\\
6638	&	\citenips{NIPS2017_6638}	&	N	&	Y	&	N	&	Y	&	NA	&	N	&	N	&	NA	&	N	&	N	\\
6651	&	\citenips{NIPS2017_6651}	&	N	&	Y	&	N	&	Y	&	NA	&	NA	&	Y	&	N	&	N	&	N	\\
6653	&	\citenips{NIPS2017_6653}	&	N	&	Y	&	N	&	N	&	N	&	Y	&	Y	&	N	&	N	&	N	\\
6659	&	\citenips{NIPS2017_6659}	&	N	&	Y	&	N	&	Y	&	NA	&	NA	&	N	&	NA	&	Y	&	N	\\
6661	&	\citenips{NIPS2017_6661}	&	N	&	Y	&	Y	&	Y	&	NA	&	NA	&	N	&	NA	&	N	&	N	\\
6676	&	\citenips{NIPS2017_6676}	&	N	&	Y	&	N	&	Y	&	NA	&	NA	&	N	&	NA	&	N	&	Y	\\
6684	&	\citenips{NIPS2017_6684}	&	N	&	Y	&	N	&	Y	&	NA	&	NA	&	N	&	NA	&	N	&	N	\\
6685	&	\citenips{NIPS2017_6685}	&	N	&	Y	&	N	&	Y	&	NA	&	NA	&	Y	&	N	&	N	&	N	\\
6691	&	\citenips{NIPS2017_6691}	&	N	&	Y	&	N	&	Y	&	NA	&	NA	&	N	&	NA	&	N	&	N	\\
6693	&	\citenips{NIPS2017_6693}	&	N	&	Y	&	N	&	Y	&	NA	&	NA	&	N	&	NA	&	N	&	Y	\\
6698	&	\citenips{NIPS2017_6698}	&	N	&	Y	&	N	&	N	&	N	&	Y	&	Y	&	N	&	Y	&	N	\\
6699	&	\citenips{NIPS2017_6699}	&	Y	&	Y	&	N	&	N	&	N	&	N	&	N	&	NA	&	N	&	N	\\
6700&	\citenips{NIPS2017_6700}	&	Y	&	Y	&	N	&	Y	&	NA	&	NA	&	N	&	NA	&	N	&	Y	\\
6701	&	\citenips{NIPS2017_6701}	&	N	&	Y	&	N	&	N	&	N	&	Y	&	Y	&	N	&	N	&	N	\\
6708	&	\citenips{NIPS2017_6708}	&	Y	&	Y	&	N	&	N	&	N	&	Y	&	N	&	NA	&	N	&	N	\\
6721	&	\citenips{NIPS2017_6721}	&	Y	&	Y	&	N	&	N	&	N	&	N	&	N	&	NA	&	N	&	N	\\
6737	&	\citenips{NIPS2017_6737}	&	Y	&	Y	&	N	&	Y	&	NA	&	NA	&	N	&	NA	&	N	&	N	\\
6743	&	\citenips{NIPS2017_6743}	&	Y	&	Y	&	N	&	Y	&	NA	&	NA	&	Y	&	N	&	N	&	N	\\
6748	&	\citenips{NIPS2017_6748}	&	N	&	Y	&	N	&	N	&	N	&	N	&	N	&	NA	&	N	&	N	\\
6753	&	\citenips{NIPS2017_6753}	&	N	&	Y	&	N	&	Y	&	NA	&	NA	&	N	&	NA	&	N	&	N	\\
6757	&	\citenips{NIPS2017_6757}	&	N	&	Y	&	N	&	Y	&	NA	&	NA	&	Y	&	N	&	N	&	N	\\
6761	&	\citenips{NIPS2017_6761}	&	N	&	Y	&	N	&	Y	&	NA	&	NA	&	N	&	NA	&	N	&	N	\\
6762	&	\citenips{NIPS2017_6762}	&	Y	&	Y	&	N	&	Y	&	NA	&	NA	&	Y	&	N	&	N	&	N	\\
6767	&	\citenips{NIPS2017_6767}	&	Y	&	Y	&	N	&	N	&	N	&	Y	&	N	&	NA	&	N	&	N	\\
6768	&	\citenips{NIPS2017_6768}	&	N	&	Y	&	N	&	Y	&	NA	&	NA	&	N	&	NA	&	N	&	N	\\
6769	&	\citenips{NIPS2017_6769}	&	N	&	Y	&	N	&	Y	&	NA	&	NA	&	N	&	NA	&	N	&	N	\\
6770	&	\citenips{NIPS2017_6770}	&	N	&	Y	&	N	&	Y	&	NA	&	NA	&	N	&	NA	&	N	&	Y	\\
6788	&	\citenips{NIPS2017_6788}	&	N	&	Y	&	N	&	N	&	N	&	N	&	N	&	NA	&	N	&	N	\\
6790	&	\citenips{NIPS2017_6790}	&	N	&	Y	&	N	&	Y	&	NA	&	NA	&	N	&	NA	&	N	&	N	\\
6791	&	\citenips{NIPS2017_6791}	&	N	&	Y	&	N	&	Y	&	NA	&	NA	&	N	&	NA	&	N	&	N	\\
6794	&	\citenips{NIPS2017_6794}	&	Y	&	Y	&	N	&	Y	&	N	&	NA	&	Y	&	N	&	N	&	Y	\\
6801	&	\citenips{NIPS2017_6801}	&	N	&	Y	&	N	&	N	&	N	&	Y	&	N	&	NA	&	N	&	N	\\
6806	&	\citenips{NIPS2017_6806}	&	Y	&	N	&	N	&	Y	&	NA	&	NA	&	N	&	NA	&	N	&	N	\\
6811	&	\citenips{NIPS2017_6811}	&	N	&	Y	&	N	&	Y	&	NA	&	NA	&	N	&	NA	&	N	&	N	\\
6813	&	\citenips{NIPS2017_6813}	&	N	&	Y	&	N	&	Y	&	NA	&	NA	&	N	&	NA	&	N	&	N	\\
6816	&	\citenips{NIPS2017_6816}	&	N	&	Y	&	N	&	Y	&	NA	&	NA	&	N	&	NA	&	N	&	N	\\
6819	&	\citenips{NIPS2017_6819}	&	N	&	Y	&	N	&	N	&	N	&	Y	&	Y	&	N	&	N	&	N	\\
6820	&	\citenips{NIPS2017_6820}	&	N	&	Y	&	N	&	Y	&	NA	&	NA	&	Y	&	N	&	N	&	N	\\
6821	&	\citenips{NIPS2017_6821}	&	N	&	Y	&	N	&	Y	&	NA	&	NA	&	N	&	NA	&	N	&	N	\\
6829	&	\citenips{NIPS2017_6829}	&	N	&	Y	&	N	&	Y	&	NA	&	NA	&	N	&	NA	&	N	&	N	\\
6835	&	\citenips{NIPS2017_6835}	&	Y	&	Y	&	N	&	N	&	N	&	Y	&	N	&	NA	&	N	&	N	\\
6838	&	\citenips{NIPS2017_6838}	&	N	&	Y	&	Y	&	N	&	N	&	Y	&	N	&	NA	&	N	&	N	\\
6839	&	\citenips{NIPS2017_6839}	&	Y	&	Y	&	N	&	N	&	N	&	Y	&	Y	&	N	&	N	&	N	\\
6842	&	\citenips{NIPS2017_6842}	&	N	&	Y	&	N	&	N	&	N	&	Y	&	N	&	NA	&	N	&	N	\\
6854	&	\citenips{NIPS2017_6854}	&	N	&	Y	&	N	&	Y	&	NA	&	NA	&	N	&	NA	&	N	&	N	\\
6862	&	\citenips{NIPS2017_6862}	&	N	&	Y	&	N	&	N	&	N	&	Y	&	N	&	NA	&	N	&	N	\\
6864	&	\citenips{NIPS2017_6864}	&	N	&	Y	&	N	&	N	&	N	&	Y	&	N	&	NA	&	N	&	N	\\
6866	&	\citenips{NIPS2017_6866}	&	Y	&	Y	&	N	&	N	&	N	&	Y	&	Y	&	N	&	N	&	N	\\
6871	&	\citenips{NIPS2017_6871}	&	N	&	Y	&	Y	&	Y	&	NA	&	NA	&	N	&	NA	&	N	&	N	\\
6872	&	\citenips{NIPS2017_6872}	&	N	&	Y	&	Y	&	Y	&	NA	&	N	&	N	&	NA	&	N	&	N	\\
6877	&	\citenips{NIPS2017_6877}	&	Y	&	Y	&	N	&	N	&	N	&	N	&	N	&	NA	&	N	&	N	\\
6879	&	\citenips{NIPS2017_6879}	&	N	&	Y	&	N	&	N	&	N	&	N	&	N	&	NA	&	N	&	N	\\
6886	&	\citenips{NIPS2017_6886}	&	N	&	Y	&	N	&	N	&	N	&	Y	&	Y	&	N	&	N	&	N	\\
6890	&	\citenips{NIPS2017_6890}	&	N	&	Y	&	N	&	N	&	N	&	N	&	Y	&	N	&	N	&	N	\\
6892	&	\citenips{NIPS2017_6892}	&	N	&	Y	&	N	&	N	&	N	&	Y	&	N	&	NA	&	N	&	N	\\
6907	&	\citenips{NIPS2017_6907}	&	N	&	Y	&	N	&	N	&	NA	&	N	&	N	&	NA	&	N	&	N	\\
6916	&	\citenips{NIPS2017_6916}	&	N	&	Y	&	N	&	N	&	N	&	Y	&	Y	&	N	&	N	&	N	\\
6927	&	\citenips{NIPS2017_6927}	&	N	&	Y	&	N	&	N	&	N	&	N	&	N	&	NA	&	N	&	N	\\
6931	&	\citenips{NIPS2017_6931}	&	Y	&	Y	&	Y	&	Y	&	NA	&	NA	&	N	&	NA	&	N	&	N	\\
6933	&	\citenips{NIPS2017_6933}	&	Y	&	Y	&	N	&	N	&	N	&	Y	&	Y	&	N	&	N	&	N	\\
6934	&	\citenips{NIPS2017_6934}	&	N	&	Y	&	N	&	Y	&	NA	&	NA	&	Y	&	N	&	N	&	N	\\
6936	&	\citenips{NIPS2017_6936}	&	N	&	Y	&	N	&	Y	&	NA	&	NA	&	N	&	NA	&	N	&	N	\\
6937	&	\citenips{NIPS2017_6937}	&	N	&	Y	&	N	&	N	&	N	&	Y	&	Y	&	N	&	N	&	N	\\
6946	&	\citenips{NIPS2017_6946}	&	Y	&	Y	&	N	&	Y	&	NA	&	NA	&	N	&	NA	&	N	&	N	\\
6952	&	\citenips{NIPS2017_6952}	&	N	&	Y	&	N	&	Y	&	NA	&	NA	&	N	&	NA	&	N	&	N	\\
6960	&	\citenips{NIPS2017_6960}	&	Y	&	Y	&	N	&	N	&	N	&	Y	&	N	&	NA	&	N	&	N	\\
6963	&	\citenips{NIPS2017_6963}	&	N	&	Y	&	Y	&	Y	&	NA	&	NA	&	Y	&	N	&	N	&	Y	\\
6964	&	\citenips{NIPS2017_6964}	&	N	&	Y	&	N	&	Y	&	NA	&	NA	&	Y	&	N	&	N	&	N	\\
6966	&	\citenips{NIPS2017_6966}	&	N	&	Y	&	N	&	Y	&	NA	&	NA	&	N	&	NA	&	N	&	N	\\
6976	&	\citenips{NIPS2017_6976}	&	N	&	Y	&	N	&	Y	&	NA	&	NA	&	N	&	NA	&	N	&	N	\\
6978	&	\citenips{NIPS2017_6978}	&	N	&	Y	&	N	&	Y	&	NA	&	NA	&	N	&	NA	&	NA	&	N	\\
6979	&	\citenips{NIPS2017_6979}	&	N	&	Y	&	N	&	Y	&	NA	&	NA	&	Y	&	N	&	N	&	N	\\
6984	&	\citenips{NIPS2017_6984}	&	N	&	Y	&	N	&	N	&	N	&	Y	&	N	&	NA	&	N	&	N	\\
6988	&	\citenips{NIPS2017_6988}	&	N	&	Y	&	N	&	N	&	N	&	Y	&	Y	&	N	&	N	&	N	\\
6998	&	\citenips{NIPS2017_6998}	&	Y	&	Y	&	Y	&	N	&	N	&	Y	&	N	&	NA	&	N	&	N	\\
6999	&	\citenips{NIPS2017_6999}	&	Y	&	Y	&	N	&	N	&	N	&	N	&	Y	&	N	&	Y	&	N	\\
7004	&	\citenips{NIPS2017_7004}	&	N	&	Y	&	N	&	N	&	N	&	Y	&	N	&	NA	&	N	&	N	\\
7022	&	\citenips{NIPS2017_7022}	&	Y	&	Y	&	Y	&	Y	&	NA	&	NA	&	N	&	NA	&	N	&	N	\\
7033	&	\citenips{NIPS2017_7033}	&	N	&	Y	&	N	&	Y	&	NA	&	NA	&	N	&	NA	&	N	&	N	\\
7045	&	\citenips{NIPS2017_7045}	&	N	&	Y	&	N	&	Y	&	NA	&	NA	&	N	&	NA	&	N	&	N	\\
7047	&	\citenips{NIPS2017_7047}	&	Y	&	Y	&	N	&	N	&	N	&	N	&	N	&	NA	&	N	&	N	\\
7048	&	\citenips{NIPS2017_7048}	&	Y	&	Y	&	N	&	Y	&	NA	&	NA	&	Y	&	N	&	N	&	N	\\
7053	&	\citenips{NIPS2017_7053}	&	N	&	Y	&	N	&	N	&	N	&	Y	&	N	&	NA	&	N	&	N	\\
7056	&	\citenips{NIPS2017_7056}	&	N	&	Y	&	N	&	N	&	N	&	Y	&	N	&	NA	&	N	&	N	\\
7058	&	\citenips{NIPS2017_7058}	&	Y	&	Y	&	N	&	Y	&	NA	&	NA	&	N	&	NA	&	N	&	N	\\
7064	&	\citenips{NIPS2017_7064}	&	N	&	Y	&	N	&	N	&	N	&	N	&	N	&	NA	&	N	&	N	\\
7071	&	\citenips{NIPS2017_7071}	&	N	&	Y	&	N	&	N	&	N	&	Y	&	N	&	NA	&	N	&	N	\\
7073	&	\citenips{NIPS2017_7073}	&	N	&	Y	&	N	&	N	&	N	&	N	&	N	&	NA	&	N	&	N	\\
7076	&	\citenips{NIPS2017_7076}	&	Y	&	Y	&	N	&	N	&	N	&	N	&	N	&	NA	&	N	&	N	\\
7080	&	\citenips{NIPS2017_7080}	&	Y	&	Y	&	Y	&	Y	&	NA	&	NA	&	N	&	NA	&	N	&	N	\\
7081	&	\citenips{NIPS2017_7081}	&	N	&	Y	&	N	&	Y	&	NA	&	NA	&	N	&	NA	&	N	&	Y	\\
7093	&	\citenips{NIPS2017_7093}	&	Y	&	Y	&	N	&	N	&	Y	&	Y	&	N	&	NA	&	N	&	N	\\
7098	&	\citenips{NIPS2017_7098}	&	Y	&	Y	&	N	&	N	&	N	&	Y	&	Y	&	N	&	N	&	N	\\
7102	&	\citenips{NIPS2017_7102}	&	N	&	Y	&	N	&	N	&	N	&	Y	&	N	&	NA	&	N	&	N	\\
7103	&	\citenips{NIPS2017_7103}	&	N	&	Y	&	N	&	Y	&	NA	&	NA	&	N	&	NA	&	N	&	N	\\
7107	&	\citenips{NIPS2017_7107}	&	N	&	Y	&	N	&	Y	&	NA	&	NA	&	N	&	NA	&	N	&	N	\\
7110	&	\citenips{NIPS2017_7110}	&	N	&	Y	&	N	&	Y	&	NA	&	NA	&	N	&	NA	&	N	&	N	\\
7111	&	\citenips{NIPS2017_7111}	&	Y	&	Y	&	N	&	Y	&	NA	&	NA	&	Y	&	N	&	N	&	N	\\
7125	&	\citenips{NIPS2017_7125}	&	N	&	Y	&	N	&	N	&	N	&	Y	&	N	&	NA	&	N	&	N	\\
7133	&	\citenips{NIPS2017_7133}	&	Y	&	Y	&	N	&	Y	&	NA	&	NA	&	Y	&	N	&	N	&	N	\\
7147	&	\citenips{NIPS2017_7147}	&	N	&	Y	&	N	&	Y	&	NA	&	NA	&	Y	&	N	&	N	&	N	\\
7170	&	\citenips{NIPS2017_7170}	&	N	&	Y	&	N	&	N	&	N	&	Y	&	Y	&	N	&	N	&	N	\\
7172	&	\citenips{NIPS2017_7172}	&	N	&	Y	&	N	&	N	&	N	&	Y	&	N	&	NA	&	N	&	N	\\
7182	&	\citenips{NIPS2017_7182}	&	N	&	Y	&	N	&	Y	&	NA	&	NA	&	Y	&	N	&	N	&	N	\\
7191	&	\citenips{NIPS2017_7191}	&	N	&	Y	&	N	&	N	&	Y	&	Y	&	Y	&	N	&	N	&	N	\\
7219	&	\citenips{NIPS2017_7219}	&	Y	&	Y	&	N	&	Y	&	NA	&	NA	&	Y	&	N	&	N	&	N	\\
7220	&	\citenips{NIPS2017_7220}	&	N	&	Y	&	Y	&	N	&	N	&	Y	&	N	&	NA	&	N	&	N	\\
7225	&	\citenips{NIPS2017_7225}	&	N	&	Y	&	N	&	N	&	N	&	Y	&	N	&	NA	&	N	&	N	\\
7231	&	\citenips{NIPS2017_7231}	&	N	&	Y	&	N	&	N	&	N	&	Y	&	Y	&	N	&	N	&	N	\\
7232	&	\citenips{NIPS2017_7232}	&	N	&	Y	&	N	&	Y	&	NA	&	NA	&	N	&	NA	&	N	&	N	\\
7244	&	\citenips{NIPS2017_7244}	&	N	&	Y	&	N	&	N	&	N	&	Y	&	Y	&	N	&	N	&	N	\\
7245	&	\citenips{NIPS2017_7245}	&	Y	&	Y	&	N	&	N	&	N	&	Y	&	Y	&	N	&	Y	&	N	\\
7254	&	\citenips{NIPS2017_7254}	&	N	&	Y	&	N	&	Y	&	NA	&	NA	&	N	&	NA	&	N	&	N	\\
7269	&	\citenips{NIPS2017_7269}	&	N	&	Y	&	N	&	N	&	N	&	N	&	N	&	NA	&	N	&	N	\\
7278	&	\citenips{NIPS2017_7278}	&	N	&	Y	&	N	&	Y	&	NA	&	NA	&	N	&	NA	&	N	&	N	\\
\caption{The consensus results for the 121 manuscripts for which the reviewers could agree on the outcome for each review item}
\label{tab:conresults}
\end{longtable}

\subsection{Summary of consensus classification}

\begin{figure}[h!]
\centering
\includegraphics[scale=0.4]{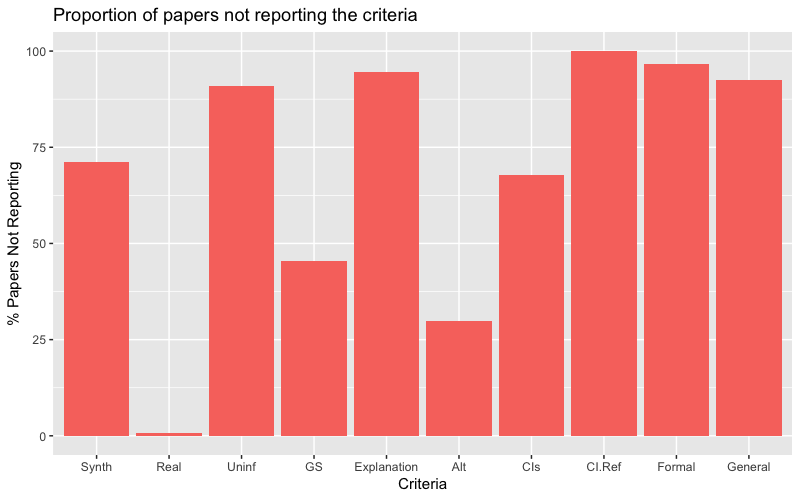}
\caption{Each bar shows the proportion of papers that did not report the criterion. Note that the sequential nature of the items means that papers are excluded dependent on previous answers. For example, the bar for "CI.Ref" is 100\%, which means that none of the papers reported a reference for how their confidence intervals were calculated. NAs occur when: Gold Standard (GS) is reported, therefore no explanation or alternative (Alt) is required and hence these are NA. Similarly if confidence intervals (CIs) are not reported then the reference for CIs are not applicable. As above the criterion can be split into three groups: Experiments, concerning which data are used to test models, Baselines, concerning if uninformed or gold-standard models are used for benchmarking and quantification, concerning reporting of uncertainty and formal comparison of model performance.}
\label{fig:qsummaries}
\end{figure}

The proportion of papers that did not report each of the criterion of interest are summarised in figure \ref{fig:qsummaries}. These bar plots should be interpreted carefully as the items are answered sequentially and papers have been excluded depending on previous answers. The tabular version of these results, along with the number of papers included in the analysis, are in Table \ref{tab:qsummaries}; the most important points to note are:
\begin{itemize}
\item Out of all eligible papers, only 3\% undertook formal hypothesis testing to compare their novel algorithm to an appropriate baseline, and only 32\% calculated confidence intervals for their reported performance metrics.
\item Of the papers that reported confidence intervals, not a single one reported a reference for, or derivation of, the maths behind these confidence intervals.
\item Nearly all papers used real-world data to test their models apart from one which did not. In contrast only 29\% used synthetic data.
\end{itemize}

\begin{table}[h!]
\centering
\begin{tabular}{||c|l||c|c||c||}
\hline\hline
Item Group & Item & Reported:Y & Reported:N & sample \\
\hline
\multirow{2}{*}{Experiments} & (1) Synth &	35(28.93)	 &86(71.07)	 &121 \\
&(2) Real &	120(99.17)	 &1(0.83)	 &121 \\
\hline
\multirow{4}{*}{Baselines}  & (3) Uninf &	11(9.09)	 &110(90.91)	 &121 \\
&(4) GS &	66(54.55)	 &55(45.45)	 &121 \\
&(4.1) Expl  &3(5.45)	 &52(94.55) &	55 \\
& (4.2) Alt &	40(70.18)	 &17(29.82)	 &57 \\
\hline
\multirow{5}{*}{Quantification}  & (5) CIs &	39(32.23)	 &82(67.77) &	121 \\
&(5.1) CI.Ref &	0(0.00)	 &39(100.00)	 &39 \\
&(6) Formal	 &4(3.33) &	116(96.67)	 &120 \\
&(7) General	 &9(7.44)	 &112(92.56)	 &121 \\
\hline\hline
\end{tabular}
\caption{Number (and proportion) of papers that reported/didn't report according to review items. Rows are criteria. First two columns (item group, item) indicate the review item, entries as described in Figure~\ref{fig:questions} and discussed in Section~\ref{Sec:methods.protocol} (in same sequence). Second two columns (Reported:Y, Reported:N) contain the absolute counts, and in brackets percentage frequencies relative to the applicable sample; entries in the third column (Reported:Y) indicate how many papers out of the sample did report, fourth column (Reported:N) indicates how many papers out of the sample did not report the item corresponding to the row. Fifth column (sample) indicates the size of the applicable sample out of a total of 121 papers, which may be reduced through conditionalities in items as outlined in Section~\ref{Sec:methods.protocol}.}
\label{tab:qsummaries}
\end{table}

\newpage

\subsection{Inter-reviewer agreement: summaries and exploration}

Agreement between the reviewer pairs is computed based on intra-paper and intra-item inter-reviewer agreement samples, one each for each reviewer pair. The intra-paper inter-reviewer agreement sample for a given pair AB is the sample of the N percentages of how many out of the 10 items in each of the N jointly reviewed papers that A and B did agree upon. The intra-item inter-reviewer agreement sample for a given pair AB is the sample of the 10 percentages of how many out of the N jointly reviewed papers the reviewers A and B agreed upon, for each of the 10 items. For all percentages, an agreement is considered as either both reviewers recording ``yes'', both recording ``no'', or both recording that the item does not apply to the paper.

\begin{figure}[H]
\centering
\includegraphics[scale=0.5]{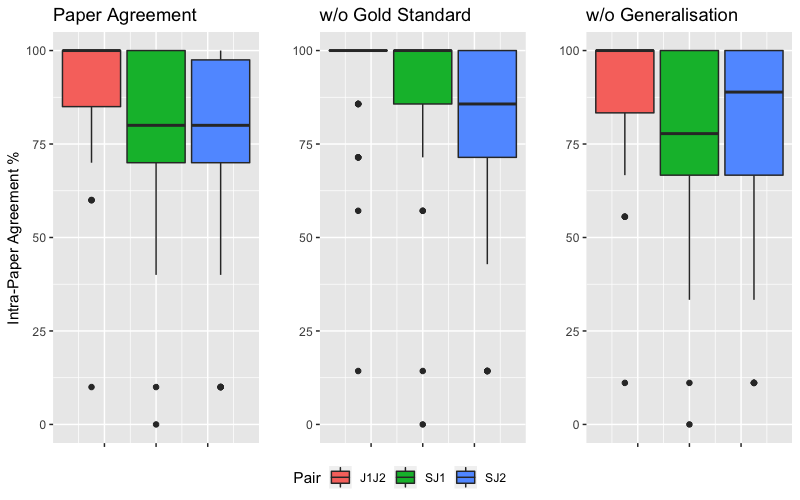}
\caption{y-axis is proportion of intra-paper agreement, x-axis within each box plot is reviewer pairing, indicating one of the three reviewer pairs SJ1, SJ2, and J1J2. The boxplots show the spread of the intra-paper agreement across all papers the pair on the x-axis had jointly reviewed. Left: agreement computed on all 10 items. Middle: agreement computed from 7 items = 10 items excluding the three items (4) Gold Standard, (4.1) Explanation and (4.2) Alternative. Right: agreement computed from 9 items = 10 items excluding the item (7) Generalization.}
\label{fig:paperagreement}
\end{figure}

\begin{figure}[h!]
\centering
\includegraphics[scale=0.5]{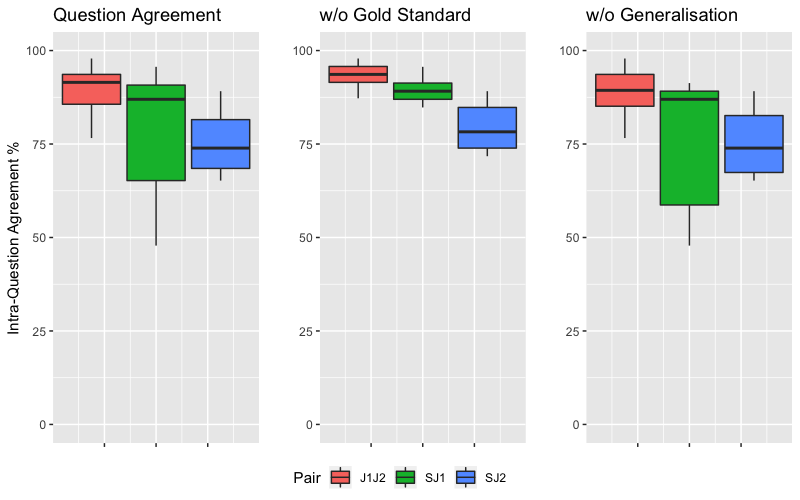}
\caption{y-axis is proportion of intra-question agreement, x-axis within each box plot is reviewer pairing, indicating one of the three reviewer pairs SJ1, SJ2, and J1J2. The boxplots show the distribution of the intra-question agreement across each of the 10 items. Left: sample of all 10 items. Middle: samples of 7 items = 10 items excluding the three items (4) Gold Standard, (4.1) Explanation and (4.2) Alternative. Right: samples of 9 items = 10 items excluding the item (7) Generalization.}
\label{fig:questionagreement}
\end{figure}

The inter-reviewer agreement samples for the three reviewer pairs are summarised below.
Table~\ref{tab:paperagreement} and Figure~\ref{fig:paperagreement} summarize the inter-paper inter-reviewer agreement sample. Table~\ref{tab:questionagreement} and Figure~\ref{fig:questionagreement} summarize the intra-item inter-reviewer agreement sample.

Tables~\ref{tab:paperagreement} and~\ref{tab:questionagreement} are six-number summaries, Figures~\ref{fig:paperagreement} and~\ref{fig:paperagreement} are box plots of the respective sample, by reviewer pair.

In Figures~\ref{fig:paperagreement} and~\ref{fig:paperagreement}, variants of the full sample box plot (left) are also included, considering agreements and samples excluding questions about the gold standard (middle) and generalization statement (right). This is to obtain a visual indication of the disagreement which these specific questions (4 and 7) introduce, based on qualitative observations during the process which led to the hypothesis that these are questions the reviewers frequently disagreed upon.
Visually, one may observe a decrease in inter-quartile range when excluding question (4), but not when excluding question (7). This indicates that there could be a high amount of disagreement about the nature of a gold standard or the state-of-art. We will revisit this qualitative observation more quantitatively in Section~\ref{Sec:results.kappas}.

\begin{table}[h!]
\centering
\begin{tabular}{|c|c|c|c|c|c|c|c|}
\hline
& Min. & Q1 & Med. & Mean & Q3 & Max. & n \\
\hline
SJ1 & 0 &	70.00	 & 80.00 & 78.91 & 100.00	 & 100.00 & 46 \\
SJ2	& 10.00 &	 70.00 & 80.00 & 75.43 & 97.50 &	100.00 & 47\\
J1J2 &	10.00 &	85.00	& 100.00 & 89.79 & 100.00 & 100.00 & 46\\
\hline
\end{tabular}
\caption{Six-number summaries of the intra-paper inter-reviewer agreement sample, for each of the three reviewer pairs SJ1, SJ2, and J1J2.}
\label{tab:paperagreement}
\end{table}

\begin{table}[h!]
\centering
\begin{tabular}{|c|c|c|c|c|c|c|c|}
\hline
& Min. & Q1 & Med. & Mean & Q3 & Max. & n \\
\hline
SJ1 & 47.83 & 65.22	& 86.96 & 78.91 & 90.76 & 95.65 & 10\\
SJ2 & 65.22	& 68.48 & 73.91 & 75.44 & 81.52	& 89.13 & 10 \\
J1J2	 & 76.60	& 85.64 &	 91.49 & 89.79	& 93.62 & 97.87 & 10\\
\hline
\end{tabular}
\caption{Six-number summaries of the intra-item inter-reviewer agreement sample, for each of the three reviewer pairs SJ1, SJ2, and J1J2.}
\label{tab:questionagreement}
\end{table}

\subsection{Inter-reviewer agreement: quantification}
\label{Sec:results.kappas}

\begin{figure}[H]
\centering
\includegraphics[scale=0.6]{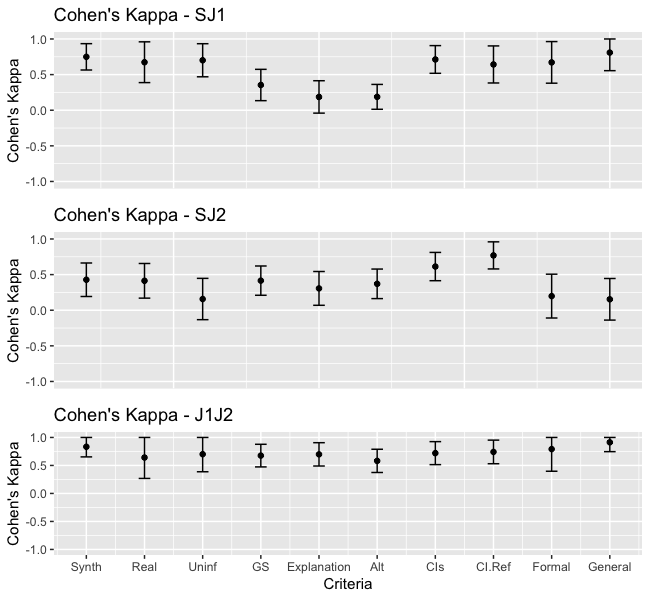}
\caption{Cohen's Kappa estimates with error bars showing agreement among reviewer pairs for each criterion. The x-axis is the criteria, the y-axis the Kappa values and the error bars give 95\% confidence intervals for the Kappas.}
\label{fig:Kappas}
\end{figure}

Figure~\ref{fig:Kappas} presents Cohen's Kappa for each question, across all reviewer pairs. The error bars are 95\% confidence intervals. To support these we also calculated Fleiss' Kappa (Table \ref{tab:fleiss}). The plot and table indicate generally excellent agreement for the SJ1 pair, except in the `Baselines' group of questions where this fell to poor, similarly for the J1J2 pair. The SJ2 pair had the lowest agreement throughout the review but still their disagreement was highest, and significantly so, in the `Baselines' group, whereas other low points of agreement were non-significant.

\begin{table}[H]
\centering
\begin{tabular}{c|c|c|c}
\hline\hline
Question & K - S1J1  & K - S1J2 & K - J1J2\\
\hline
Synthetic	& 0.75(2.2e-11)	& 0.42(5e-04)	& 0.83(1.3e-10)  \\
Real.World	& 0.67(3.3e-07)& 0.4(6.4e-04)	& 0.64(5.1e-08)\\
Uninformed	& 0.7(7.5e-10)	& 0.14(0.23)	& 0.7(6.2e-09)  \\
Gold.Standard	& 0.32(5.4e-03)	& 0.39(1.3e-03)	& 0.67(4.4e-07)   \\
Explanation	& 0.13(0.38)	& 0.29(3.2e-02) 	& 0.7(3.5e-07)  \\
Alternative	& 0.093(0.39)	& 0.35(3e-03) & 0.58(4.7e-07) \\
Confidence.Intervals	& 0.71(2.1e-10)	& 0.61(7.3e-07)	& 0.72(5.2e-08)  \\
CI.Reference	& 0.64(1.4e-05)& 0.77(1.8e-07) & 0.74(3.9e-07)  \\
Formal.Comparison	& 0.67(3.3e-07)& 0.19(0.14)  & 0.79(8e-12) \\
Generalisation.statement	& 0.81(4.2e-08)& 0.14(0.23)  & 0.91(1e-13)\\
\hline
\end{tabular}
\caption{Fleiss' Kappa, K, to analyse agreement between reviewer pairs. The p-values in parentheses are for the null hypothesis: raters are in agreement by chance; low values indicate agreement is not by chance. Kappas above 0.75 are considered `excellent' and below 0.40 are `poor'.}
\label{tab:fleiss}
\end{table}

\subsection{Papers without Consensus}
Finally, after reviewing all the eligible papers, two remained for which consensus could not be reached, even after a third reviewer provided their assessment. On closer examination, complete agreement could be reached for every criterion except for the three questions regarding the use of a `gold-standard' comparator. An extract of the reviewers' answers for these questions is in Table \ref{tab:disagreeQs}. It appears from the table that the problem stems from: firstly, an inability to agree on if a gold-standard is reported; and secondly, whether a suitable alternative is reported (in the case that no gold-standard was reported). Discussion pertaining to why the reviewers were not able to reach consensus is included later.

\begin{table}[h!]
\centering
\begin{tabular}{|c|c|c|c|c|}
\hline
Paper & Reviewer & Gold Standard & Explanation & Alternative \\
\hline
\multirow{3}{*}{6891} & S & N & 	N &	N \\
& J1 & N&	N	&Y \\
& J2 & Y & NA & NA \\
\hline
\multirow{3}{*}{6968} & S & N & 	N &	N \\
& J1 & Y & NA & NA \\
& J2 & N & N & Y \\
\hline
\end{tabular}
\caption{The criterion for which no consensus was reached. Only the `Alternative' could reach no agreement but this was due to the responses in the previous two. For example if a reviewer thought `Gold Standard' should be `Y' then the next two must be `NA'.}
\label{tab:disagreeQs}
\end{table}

\newpage
\section{Discussion}

To the best of our knowledge, our review is the first of its kind in the ML/AI domain. Whilst previous work has highlighted the lack of engagement with specific parts of the argumentative chain for empirically evidencing algorithmic effectiveness, no paper in ML/AI domain has previously sought to illustrate the completeness of these arguments in a representative corpus of the supervised learning literature.

In summary, the key findings of this review are:
\begin{enumerate}
\itemsep-0.2em
\item[(a)] Whilst almost all papers (except one) reported empirical comparisons using real-world data, it was substantially less common to see the use of synthetic data sets.
\item[(b)] A vast majority of manuscripts did not report an uninformed (random guess surrogate) baseline (91\%), and only a slight majority (55\%) reported a state-of-art baseline's performance for comparison. Though on the status of state-of-art baselines, there was less reviewer agreement; although agreement on the reporting of state-of-art contenders was significantly better than random (as measured by Kappas), it was worse than for the other items (between `good' or `poor').
\item[(c)] Of all manuscripts with reviewer consensus, about a third (32\%) reported confidence intervals for performances, while almost none (3\%) reported formal comparison quantifiers such as hypothesis testing results. The most striking finding is that none of the manuscripts reporting comparison quantifiers reported (by reference or explicitly) how exactly the quantifiers were computed (e.g., which hypothesis test was used? and how CIs were obtained?)
\end{enumerate}

\subsection{Interpretation of the Results and Strengths and Limitations of evidence}

We interpret our observations as follows:
\begin{enumerate}
\itemsep-0.2em
\item[(a)] One may argue that there is overwhelming community consensus on the scientific need for formal and quantitative model assessment, as evidenced by almost all papers reporting empirical experiments which are of benchmarking or comparison nature.
\item[(b.1)] Results may be argued to evidence a broad community consensus on the importance of comparison to representative baselines for the state-of-art. However, there seems to be only vague consensus of what constitutes such a representative baseline for a given purpose. This is mirrored by the reviewers' comparatively high level of disagreement of whether such a baseline was reported. Post-hoc scrutiny showed that some reviewers were more, some less lenient towards papers where they thought the reported baselines were widely accepted, but just not explicitly reported as such. It needs to be investigated in a future study how strong the results were impacted by the following: the reviewers' disagreement of what constitutes a baseline representative of the state-of-the-art; versus authors' failure to report one.
\item[(b.2)] The fact that the vast majority of papers do not report or reference an uninformed baseline is a very troubling finding. Without, the crucial empirical usefulness argument, i.e., that the method is better than an uninformed/naive guess, cannot be made - in 91\% of the cases. One may argue that the mere absence of this argument makes these methods untrustworthy. Since even if they are all shown to outperform a worse competitor, the new method and its supposed competitors could still simply all be worse than just guessing.
\item[(c)] There appears to be quite poor community consensus on whether and how to conduct formal quantitative comparisons, once competitor methods are chosen. The relatively frequent reporting of confidence intervals may be seen as the community's acknowledgement of the importance of quantitative comparison. Though the simultaneous absence of any explanation where (if presented) the claimed numbers are coming from may be seen as a testament of the community's lack of knowledge of how to compute them in a sensible manner, or a reliance on circulating code snippets and ad-hoc heuristics, rather than proper data scientific literature on the matter.
\end{enumerate}

\subsection{Comparison to previously published work in literature}

The results in this systematic review echo the findings of related previous work. For example, Demšar~\cite{demsar2006statistical} reports similarly infrequent engagement with the calculation of confidence intervals or related quantities in a corpus of works published between 1999 and 2003 in the proceedings of the International Conference on Machine Learning (ICML). In our review, approximately 30\% of papers reported confidence intervals, and in Demšar’s review the proportion over the reported 5 years period was between 19-48\% which is qualitatively consistent with our findings.
In contrast, reporting of formal quantitative model comparison appears to be have been much more common in the corpus reviewed by Demšar, where the proportion was as high as 53\%, compared to our review where only 3\% of 121 manuscripts reviewed undertook some quantitative comparison. However, it should be noted that Demšar included several non-parametric, and naive methods for model comparison in his analysis, which may explain the much higher proportion. If we restrict ourselves to only the manuscripts that undertook t-tests to compare models in Demšar’s review, the proportion of manuscripts undertaking hypothesis testing appears much more similar to ours, with Demšar’s proportion ranging from 4 to 16\%.

The two reviews in combination (Demšar's and ours) can be interpreted to mean that the incompleteness of empirical demonstrations for algorithmic effectiveness is a long standing issue, which has likely gotten worse, rather than improved over the last decade.

\subsection{Limitations of the Assessment Criteria}

We acknowledge that some will have principle-based criticisms of the assessment criteria utilised in this review, given its emphasis on formal comparative quantification. Our rationale for doing so is that comparison needs to involve an element of objective judgement to be scientific and empirical. Because otherwise, the act of declaring verification vs falsification is arbitrary, and does not comply with the central requirements of the scientific method. Furthermore, the extent to which chance could explain the observed findings needs to be quantified, otherwise it is unknown whether the findings could infact be explained by chance. Correctly applied frequentist hypothesis testing or Bayesian credibility intervals satisfy these requirements. We do not insist that these are the only tools to do so, but would like to point out that they seem to be the only ones used by the community (as encountered in our review).

The more practical limitation of the assessment criteria utilised is that there appears to be no consensus definition for what defines a (or the) gold-standard/state-of-art algorithm for a given purpose. As such, in situations where authors did not provide references to benchmark experiments that demonstrate a baseline algorithms prior superiority, it was difficult to determine the appropriate designation for a baseline comparator (i.e. state-of-art, or not). Similarly, there is no widely agreed upon guidance that recommends authors explicitly state what generalisation guarantees are provided by the empirical tests of effectiveness reported in their manuscript. Both of these criteria therefore relied heavily on the subjective judgements of the reviewer, which has manifested in our results as decreased agreement between the reviewer pairs, and two manuscripts for which consensus could not be reached.

At first glance, the issue of generalisation guarantees and appropriate labelling of baselines may not appear critically important to the uninitiated, but there is growing evidence that without the necessary best-practice guidance, the aforementioned ambiguity can be exploited. For example, in the medical prediction modelling domain, researcher have demonstrated how weak comparators are sometimes purposefully utilised (i.e. straw-man comparators) to inflate the gains associated with a novel algorithm ~\cite{Collinse3186}. Expecting researchers to be intimately familiar with all domains in which an algorithm can be applied is unreasonable, and therefore, the responsibility to justify why an existing algorithm should be considered the state-of-art baselines must rest with the manuscripts author.

\subsection{Implications for Future Research}

In essence, the results of this systematic review identify two high priority issues for future research:
\begin{enumerate}
\itemsep-0.2em
\item[(1)] The need for easily accessible, and statistically robust, code libraries that allow for comparison of algorithms (i.e. the generation of confidence intervals, hypothesis testing, etc.)l
\item[(2)] The need for consensus guidance on reporting to prevent the issues discussed above, e.g. recognition of baselines as being gold-standard/state-of-art, or not.
\end{enumerate}

\subsection{Strengths and Limitations of the Review}

The main limitation of this review is the use of a single journal as the primary information source, which brings into question whether the results presented are truly representative of the ML/AI literature as a whole. However, we would argue that the annual NeurIPS conference is widely recognised to be amongst the top tier, where it concerns machine learning and artificial intelligence research. Hence, it is not unreasonable to consider the NeurIPS corpus to be representative of high quality ML/AI research. By logical contra-position, NIPS may therefore equally be considered as a representative lower bound on the entire field, in terms of negative issues or verifiable failures. As such, whilst the generalisability of the results presented in this systematic review needs to be confirmed, we would argue that they are sufficiently robust to support the claims made above, i.e., that the ML/AI literature is far too often incomplete where it concerns argumentative completeness in empirical demonstrations of effectiveness.

One of the main strengths of this review was the highly conservative approach to abstract screening which was adopted, meaning that full-texts were retained unless there was a high degree of certainty that the manuscript was ineligible. This approach ultimately led to several manuscripts being progressed from screening to full-text review which could have been excluded earlier, but can be interpreted to mean that the likelihood of a potentially eligible manuscript being inappropriately excluded at the screening stage is very low. Furthermore, the independent assessment by a third reviewer at both the screening and full-text assessment stages, where the initial pair lacked agreement, provided another safety net to ensure that all relevant information was captured in the review process.

\subsection{Conclusion}

	Using a contemporary sample (2017) of research from the NeurIPS (n\'ee NIPS) supervised-learning corpus as a conservative indicator for the general quality and trustworthiness of typical contemporary ML/AI research, it would appear as though full argumentative chains in demonstrations of algorithmic effectiveness are rare. More precisely, in all the publications at NIPS 2017 which reported a new supervised learning methodology, there was at best 2 examples of complete scientific arguments capable of supporting a conclusion of effectiveness, i.e., an argument that is in-principle sufficient to evidence that ``the new method predicts with lower error than (suitable) state-of-art baselines'' \citenips{NIPS2017_6698,NIPS2017_7245}. There are many plausible reasons for this observation, some of which have been discussed, but in our opinion they can be summarised into a single issue: a lack of unambiguous reporting standards for supervised learning research specifically, or ML/AI research in general.

\subsection{A Way Forward?}
On the \emph{technical} side, the solution to the problem is straightforward: any and all future empirical supervised learning research should at a minimum provide clear justifications for the baselines and methods used, explicitly identify the gold-standard/state-of-art method, underpin comparisons by an appropriate significance/credibility quantifier, and state precisely the application cases for which the results provide guarantees for.\\

The \emph{sociological} and \emph{political} solution, unfortunately, seems much harder. It is common knowledge that current publication and review mechanisms of the field encourage grandiose claims (reviewers like them) while discouraging careful empirical argumentation (at best, reviewers ignore them). We do not anticipate any substantial change as long as the mechanisms remain as they are. Given that the vast majority publications in the ML/AI field suffer from empirical shortcomings (as we have shown), from a game theoretical perspective it also seems very unlikely that the field will see a change from within. Instead, in line with the suggestions made by \cite{ioannidis2015make}, a plausible remedy is end users - for instance, industrial or government decision makers - exerting pressure on an upstream field which causes damage (financially, and societally) through every single output that is ineffective, or ``not even wrong''.

\subsubsection*{Acknowledgements}
FK conceived the original idea of the review. BM designed the review protocol and methodology, and oversaw the review process. The three authors together conducted the review. RS conducted post-review statistical analyses and designed their presentation. FK substantially contributed to presentation and discussion of technical data science content. The three authors wrote, edited, and reviewed the manuscript together. They are jointly and equally responsible for its contents.	


\bibliographystyle{plain}
\bibliography{validation}

\bibliographystylenips{plain}
\bibliographynips{validation}

\end{document}